\pdfoutput=1

\documentclass[10pt,twocolumn,letterpaper]{article}
\PassOptionsToPackage{dvipsnames}{xcolor}
\usepackage{xcolor}
\usepackage{amssymb}
\usepackage{graphicx}
\usepackage{algorithm}
\usepackage{algpseudocode}
\usepackage{wrapfig}
\usepackage{caption}
\usepackage{subcaption}
\usepackage{bbm}
\usepackage{glossaries-extra}
\DeclareMathOperator*{\argmax}{arg\,max}

\usepackage{graphicx}
\usepackage{pdfpages}
\usepackage{titlesec}
\usepackage{bbm}
\usepackage{afterpage}

\usepackage[pagenumbers]{cvpr} 

\usepackage[dvipsnames]{xcolor}

\definecolor{cvprblue}{rgb}{0.21,0.49,0.74}
\usepackage[pagebackref,breaklinks,colorlinks,citecolor=cvprblue]{hyperref}

\def\paperID{166} 
\def\confName{3DV\xspace}
\def\confYear{2026\xspace}

\title{VIN-NBV: A View Introspection Network for Next-Best-View Selection}

\author{%
Noah Frahm \quad
Dongxu Zhao \thanks{Equal contribution.} \quad
Andrea Dunn Beltran \footnotemark[1]\\ \quad
Ron Alterovitz \quad
Jan-Michael Frahm \quad
Junier Oliva \quad
Roni Sengupta \\
\and
The University of North Carolina\\
}

\begin{document}

\twocolumn[{
\renewcommand\twocolumn[1][]{#1}
    \maketitle
    \begin{center}
    \captionsetup{type=figure}
    \includegraphics[width=\textwidth]{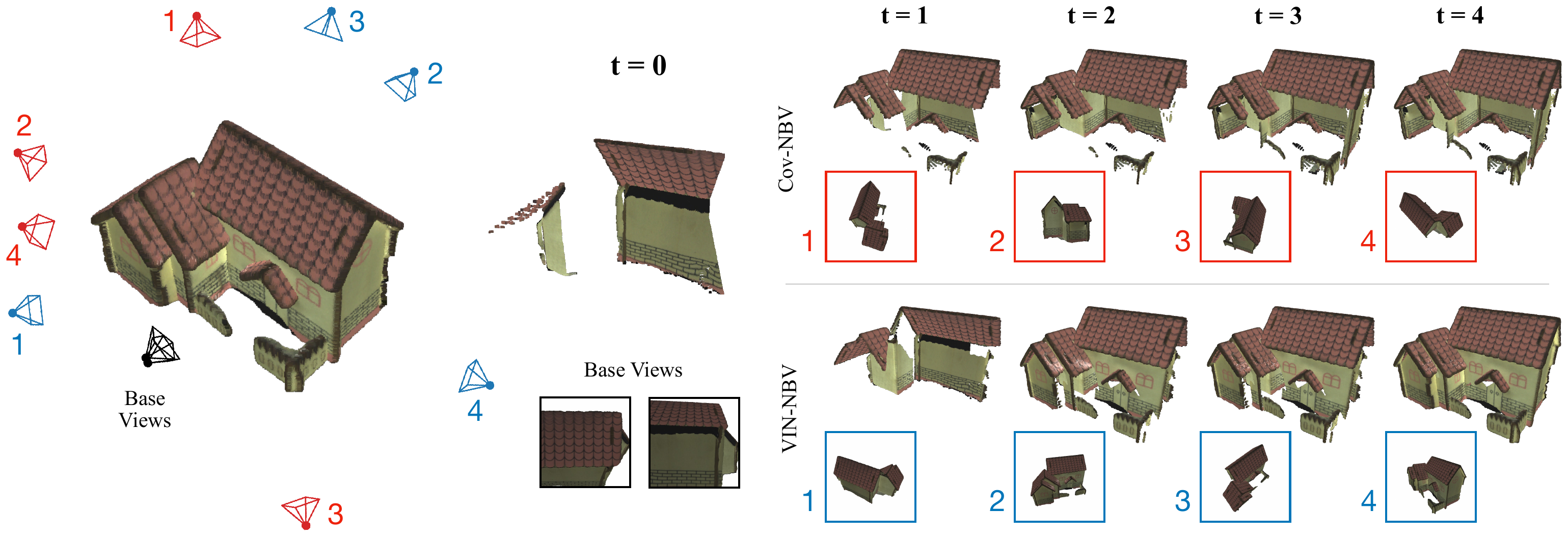}
    \captionof{figure}{
    Our approach, VIN-NBV (blue), is an NBV policy that selects best next views by maximizing a predicted reconstruction quality criterion and outperforms coverage maximization policy, Cov-NBV (red), achieving higher reconstruction accuracy with less captures.
    }
    \label{fig:teaser}
    \end{center}
}]

\begin{abstract}
\vspace{-1em}

Next Best View (NBV) algorithms aim to maximize 3D scene acquisition quality using minimal resources, e.g. number of acquisitions, time taken, or distance traversed. Prior methods often rely on coverage maximization as a proxy for reconstruction quality, but for complex scenes with occlusions and finer details, this is not always sufficient and leads to poor reconstructions. Our key insight is to train an acquisition policy that directly optimizes for reconstruction quality rather than just coverage. To achieve this, we introduce the View Introspection Network (VIN): a lightweight neural network that predicts the Relative Reconstruction Improvement (RRI) of a potential next viewpoint without making any new acquisitions. We use this network to power a simple, yet effective, sequential sampling-based greedy NBV policy. Our approach, VIN-NBV, generalizes to unseen object categories, operates without prior scene knowledge, is adaptable to resource constraints, and can handle occlusions. We show that our RRI fitness criterion leads to a $\sim$30\% gain in reconstruction quality over a coverage-based criterion using the same greedy strategy. Furthermore, VIN-NBV also outperforms deep reinforcement learning methods, Scan-RL and GenNBV, by $\sim$40\%. For more result visuals please refer to our \href{https://noahfrahm.github.io/VIN-NBVProjectPage/}{Project page.}

\end{abstract}

{\renewcommand\thefootnote{\fnsymbol{footnote}}\footnotetext[1]{These authors contributed equally.}}    
\vspace{-1.25em}
\section{Introduction}
\label{sec:intro}
\vspace{-0.25em}

Acquiring 3D knowledge of an environment is often a crucial step for many robotics applications; e.g., a drone surveying a disaster zone to assist search and rescue efforts, or autonomous robots monitoring large construction and agricultural sites. 
Real-world environments contain diverse objects of varying sizes, occlusions, and geometries; such complexities currently require a slow dense scan \cite{furukawa2010towards, schonberger2016pixelwise, frahm2010building, agarwal2011building, bleyer2011patchmatch, furukawa2009accurate, goesele2007multi, strecha2004wide} to reconstruct the 3D scene effectively. However, in many search and rescue operations, where speed leads to successful outcomes, acquiring the 3D scene in the shortest time possible is critical. In large construction or agricultural sites, battery life can be a limiting factor, making long captures costly. These challenges motivated techniques for resource-efficient scanning and 3D reconstruction, where constraints such as the number of captures, travel distance, or battery life must be respected. This problem has often been called the Next Best View (NBV) selection problem, where the goal is to predict a set of optimal camera viewpoints to maximize reconstruction quality.

Predicting NBVs is challenging due to the large search space for selecting the pose of each view, efficiently computing the optimal solution in this non-convex space, and the complexity of scene geometry and occlusions. Earlier works on NBV often assume prior knowledge of the scene (e.g., a preliminary scan or CAD model) \cite{devrim2017reinforcement,sun2021learning,zhang2021continuous,yan2021sampling, jing2016view, zhou2020offsite}, which limits applications in unexplored environments. Approaches that do not require prior knowledge of the scene predict NBVs by either maximizing coverage in the scene \cite{maver1993occlusions,roberts2017submodular,hepp2018learn,peralta2020next,guedon2022scone,guedon2023macarons} or by maximizing information gain \cite{hepp2018plan3d,zhang2021continuous,jiang2023fisherrf} in new viewpoints. While earlier methods have relied on heuristics or optimization \cite{maver1993occlusions,roberts2017submodular}, recent methods often train a deep reinforcement learning (RL) algorithm \cite{hepp2018learn,peralta2020next,hepp2018plan3d,zhang2021continuous,ran2023neurar,jiang2023fisherrf,9340916,chen2024gennbv} to maximize coverage. 

While maximizing coverage with RL \cite{chen2024gennbv,peralta2020next} may lead to a generalizable policy that can provide a good approximation of the scene, it ignores the fact that certain regions from the same scene may have more complicated geometry and self-occlusion. For example, in Fig.~\ref{fig:teaser}, if the fence of the house occludes a part of the wall, a single viewpoint capture may satisfy the coverage criterion, but will lead to a poor reconstruction with holes in the wall unless additional viewpoints are captured. Thus, our key idea is to develop an NBV acquisition policy that is trained to directly maximize the 3D reconstruction quality instead of only relying on a coverage criterion, which may fail to revisit areas with low coverage gains that will likely improve the reconstruction quality. Our policy design assumes the agent acquires RGB-D images and has no prior information about the scene.

We introduce a View Introspection Network (VIN) that can predict the relative reconstruction improvement (RRI) of any potential next view over the base images, without actually acquiring any image from the viewpoint. To demonstrate the effectiveness of directly optimizing reconstruction quality over coverage criterion, we design a simple, but effective, greedy sequential next best view selection policy VIN-NBV. At each acquisition step, the policy evaluates a set of uniformly sampled query views around the object, using the VIN to predict their effectiveness in improving reconstruction quality. The view that maximizes reconstruction improvement over base views is chosen. Our proposed VIN-NBV policy is generalizable to unseen object categories, and can operate under varying constraints, e.g. number of acquisitions, time traversed, and distance traversed; it can also avoid collisions.

The VIN is a lightweight neural network that takes the already captured base images, their camera parameters, and the query viewpoint camera parameters as input. The VIN performs a 3D-aware featurization of the already captured base views and then predicts a fitness score for each query view. We use the fitness score to indicate what Relative Reconstruction Improvement (RRI) over base views a query view may have. The VIN is trained using an imitation learning approach, where the RRI of all query viewpoints is pre-computed by explicitly reconstructing the scene with each query view and calculating the reduction in error.

We show that a relative reconstruction improvement (RRI) fitness criterion predicted by the VIN leads to ${\sim} 30\%$ improvement in reconstruction quality over a coverage fitness criterion, when using the same sampling-based greedy sequential strategy. We further show that the VIN-NBV policy significantly outperforms existing RL-based algorithms that aim to maximize coverage, reducing reconstruction error by 41\% compared to Scan-RL \cite{peralta2020next} and 39\% compared to GenNBV \cite{chen2024gennbv}.
\section{Related work}
\label{sec:related_work}
\vspace{-0.25em}

\noindent\textbf{NBV methods requiring prior knowledge.} Many existing approaches assume prior knowledge of the 3D scene, which limits applications in previously unexplored environments. Some approaches \cite{devrim2017reinforcement, sun2021learning, zhang2021continuous} directly exploit a pre-existing 3D model or its approximation. Other methods rely on 2D maps, such as estimating the height of buildings as a rough 3D model \cite{jing2016view} or obtaining a 2.5D model of the scene \cite{zhou2020offsite}.
For scenes lacking pre-existing information, often a drone fly-through along a default trajectory is made to obtain an initial coarse reconstruction \cite{roberts2017submodular,hepp2018plan3d}. In contrast, our method does not require any prior information about the scene; instead, we start with two adjacent views and sequentially choose the best next views until a specific termination criterion is met. 

\noindent\textbf{Optimal view selection with expensive scene representations.}
A related line of work addresses the problem of selecting an optimal subset of images for reconstruction by using complex scene representations. These representations are often built from existing dense capture. Although the selection algorithms themselves do not require dense pre-capture imagery as input, they rely on scene representations, such as dense point clouds, meshes, or pretrained radiance fields, that are themselves costly to obtain. Earlier approaches targeted optimal view selection for Multi-View Stereo reconstruction \citep{hornung2008image, furukawa2010towards}, while more recent work explored view selection for Neural Radiance Fields (NeRF) and its variants \citep{smith2022uncertainty, pan2022activenerf, lee2023so, jiang2023fisherrf}. In contrast, our approach does not assume access to a dense capture or an expensive prebuilt representation; we acquire images directly from the optimal viewpoints predicted by the NBV policy.

\noindent\textbf{Criteria for predicting NBV.} Previous approaches use proxy metrics like coverage \cite{maver1993occlusions, peralta2020next, drone_save, hepp2018learn, chen2024gennbv} or information gain \cite{lee2022uncertaintyguidedpolicyactive, Islerinformationgainvolumetric3d, POTTHAST2014148, jiang2023fisherrf} for predicting NBVs. State-of-the-art NBV techniques, GenNBV \cite{chen2024gennbv} and ScanRL \cite{peralta2020next}, use Reinforcement Learning (RL) with coverage-based reward functions. While coverage-based policies may lead to reasonable reconstruction quality, they fail to account for complex structures and occlusions, often resulting in holes in the reconstructed scene. On the other hand, the information gain criterion is image-based and often lacks 3D knowledge of the scene, restricting generalization. Our work goes beyond coverage by directly maximizing the reconstruction quality, which leads to significant improvement under resource constraints, as shown empirically. Rather than relying on RL \cite{peralta2020next,chen2024gennbv}, we design a greedy sequential policy, VIN-NBV, trained with imitation learning to help predict the reconstruction improvement of next views.

\begin{figure*}[!t]
  \centering
  \begin{subfigure}[b]{0.6\linewidth}
    \centering
    \includegraphics[width=\linewidth]{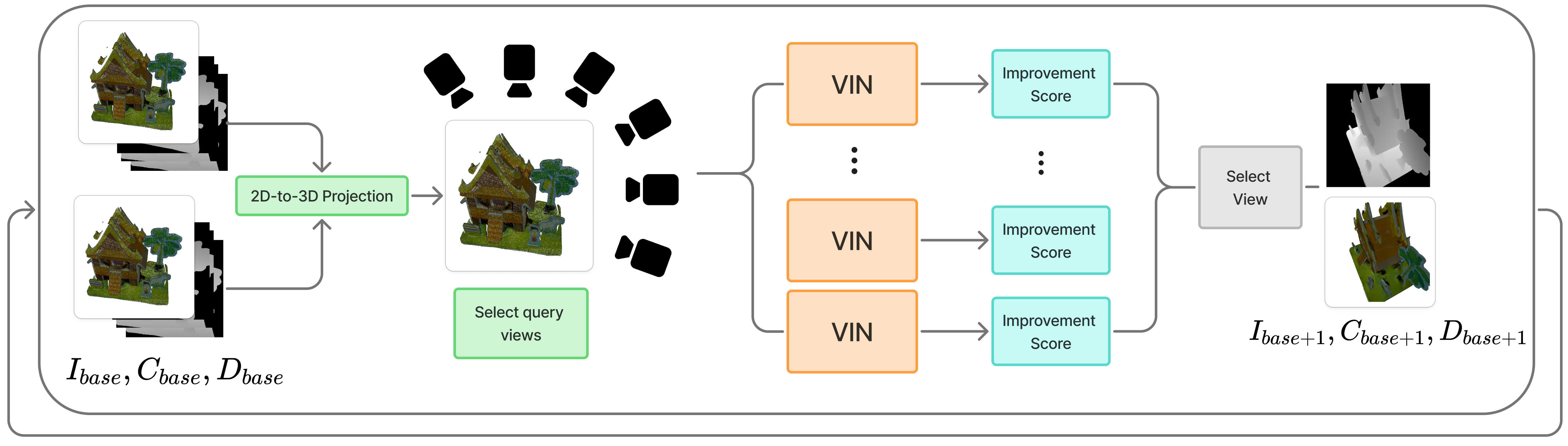}
    \caption{VIN-NBV Policy Overview}
    \label{fig:arch_left}
  \end{subfigure}
  \hfill
  \begin{subfigure}[b]{0.39\linewidth}
    \centering
    \includegraphics[width=\linewidth]{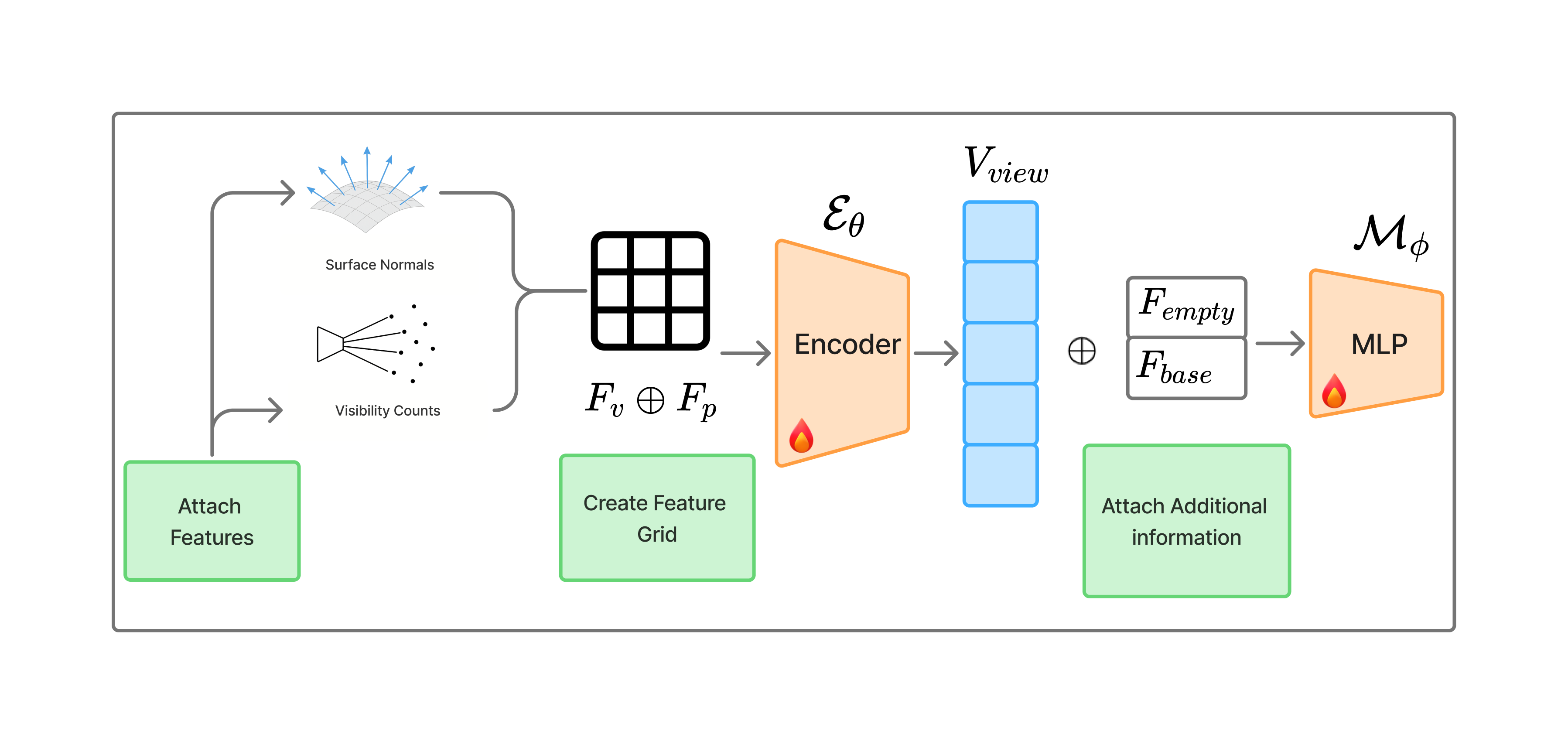}
    \caption{VIN Architecture}
    \label{fig:arch_right}
  \end{subfigure}
  \vspace{-0.5em}
  \caption{
(a) VIN-NBV reconstructs the 3D scene from prior RGB-D captures, samples candidate viewpoints, and selects the one with the highest fitness predicted by the View Introspection Network (VIN) as the Next Best View (NBV), repeating until termination.
(b) VIN uses a 3D-aware featurization, including surface normals, visibility count, and coverage ($F_{empty}$), and is trained via imitation learning to predict fitness as the Relative Reconstruction Improvement over current observations.
}
  \label{fig:method_nbv_diagram}
  \vspace{-1.2em}
\end{figure*}

\section{Method}
\label{sec:method}
\vspace{-0.25em}

In the following section, we formalize the NBV problem and describe our VIN-NBV policy in detail.
We will first provide a mathematical overview of the NBV problem in Sec. \ref{ssec:prob}. Then, we will introduce our proposed sampling-based greedy NBV policy, VIN-NBV, in Sec. \ref{ssec:policy}, followed by the design of the View Introspection Network (VIN)  in \ref{ssec:vin_design} and its training in \ref{ssec:vin_training}.

\subsection{Problem Setup}
\label{ssec:prob}
\vspace{-0.25em}

Consider an agent that has acquired $k$ initial base images of a scene $I_{base}=\{I_{1},...,I_{k}\}$ from viewpoints with camera parameters $C_{base}=\{C_{1},...,C_{k}\}$ and depth maps $D_{base}=\{D_{1},...,D_{k}\}$ either captured with a depth sensor or predicted with a Monocular Depth Estimator or Multi-View Stereo algorithms. The agent then runs a 3D reconstruction pipeline using the initial base views to reconstruct the 3D scene as $\mathcal{R}_{base}$. The goal of NBV is to predict a set of $m$ next best views $C_{nbv}=\{C_{{1}},...,C_{m}\}$ from which the scene should be captured to maximize the reconstruction quality of the scene. 

More specifically, from $C_{nbv}$ we capture NBV images $I_{nbv}=\{I_{1},...,I_{m}\}$ and associated depth maps $D_{nbv}=\{D_{1},...,D_{m}\}$ and perform 3D reconstruction to create $\mathcal{R}_{final}$ using $I_{base} \cup I_{nbv}$ and $D_{base} \cup D_{nbv}$. While previous NBV techniques maximize coverage, we maximize reconstruction quality measured using the relative improvement of the Chamfer Distance of $\mathcal{R}_{final}$ over $\mathcal{R}_{base}$:
\begin{equation}
     C^{*}_{nbv} = \argmax_{C_{nbv}} ~\frac{CD(\mathcal{R}_{base},\mathcal{R}_{GT}) -CD(\mathcal{R}_{final},\mathcal{R}_{GT})}{CD(\mathcal{R}_{base},\mathcal{R}_{GT})},     
    \label{eq:cd-criteria}
\end{equation}

where $CD(\mathcal{R},\mathcal{R}_{GT})$ calculates the Chamfer Distance between the reconstructed point cloud of scene $\mathcal{R}$ and the ground-truth point cloud $\mathcal{R}_{GT}$.

\setlength{\intextsep}{0pt}
\setlength{\textfloatsep}{1pt}
\begin{algorithm}[!htb]
  \caption{VIN-NBV Policy}
  \label{method_nbv_algorithm}
  \begin{algorithmic}[1]
    \State $I^k_{\mathrm{base}}\gets\{I_1,\dots,I_k\}$
    \State $C^k_{\mathrm{base}}\gets\{C_1,\dots,C_k\}$
    \State $D^k_{\mathrm{base}}\gets\{D_1,\dots,D_k\}$
    \State $t = k$
    \While{not termination\_criteria()}
      \State Reconstruct $R^t_{\mathrm{base}}$ from $(I^t_{\mathrm{base}}, C^k_{\mathrm{base}}, D^k_{\mathrm{base}})$
      \State Sample a set of query views $\{q_i\}$
      \For{$i=1,\dots,n$}
        \State $\widehat{\mathcal{RRI}}(q_i) = \text{VIN}_\theta(R^{t}_{base}, C^t_{base}, C_{q_i})${\label{line:score_func}}
      \EndFor
        \State $C^t_*\gets\arg\max_q \mathcal{RRI}(q)$
        \State Capture (or render) $I^t_{*}$ and $D^{t}_{*}$ from $C^t_*$
        \State $I^{t+1}_{base} \leftarrow I^t_{base} \cup I^t_{*}$
        \State $D^{t+1}_{base} \leftarrow D^t_{base} \cup D^t_{*}$
        \State $C^{t+1}_{base} \leftarrow C^t_{base} \cup C^t_{*}$
        \State $t \mathrel{+}= 1$
    \EndWhile
    \State \Return $R_{\mathrm{final}}$
  \end{algorithmic}
\end{algorithm}

\vspace{0.5em}
\subsection{NBV Policy with View Introspection Network}
\label{ssec:policy}
\vspace{-0.25em}
We propose a simple greedy imitation learning based approach, where for each acquisition, the agent samples a set of query camera viewpoints and evaluates their fitness, and chooses the `best' one and repeats the process until some termination criterion has been reached. This acquisition policy has been described in Algorithm \ref{method_nbv_algorithm}.

The VIN-NBV policy begins with $I^k_{base}$, $C^k_{base}$, $D^k_{base}$ then iteratively selects next views until reaching a desired termination criteria, defined to fit downstream task constraints, e.g., number of image acquisitions, time traversed, battery life, etc. At each step, we update the reconstruction $R^t_{base}$ by back-projecting the RGB-D captures into 3D space. If only RGB images are available, we reconstruct $R^t_{base}$ using Multi-View Stereo or Monocular Depth Estimation. We then sample $n$ query views around this reconstruction to evaluate their fitness and choose the 'best'. 

Our key idea is the introduction of the View Introspection Network (VIN), which independently evaluates potential query views and predicts their fitness. Instead of evaluating the fitness of each view to maximize coverage, we focus on maximizing the reconstruction quality. More specifically, we define fitness criterion as the Relative Reconstruction Improvement ($\mathcal{RRI}$) over the base views by capturing the query view $q$ as defined in eq. \ref{eq:rri}. Relative Improvement is formulated to be independent of object types and scales, which otherwise affects the Chamfer Distance computation.

\vspace{-1em}
\begin{equation}
    \mathcal{RRI}(q_i) = \frac{CD(\mathcal{R}_{base},\mathcal{R}_{GT}) -CD(\mathcal{R}_{base \cup q_i},\mathcal{R}_{GT})}{CD(\mathcal{R}_{base},\mathcal{R}_{GT})}.
    \label{eq:rri}
\end{equation}

We train VIN to predict the RRI fitness criterion $\widehat{RRI}(q)$ for a query view $q$ by taking the existing reconstruction $R_{base}$, camera parameters $C_{base}$, and query view camera parameters $C_q$ as input:
\begin{equation}
    \widehat{\mathcal{RRI}}(q) = \text{VIN}_\theta(R_{base}, C_{base}, C_q).
\label{eq:vin}
\end{equation}
\vspace{-1.7em}

The design of VIN's neural architecture and its training are described in Sec \ref{ssec:vin_design} and \ref{ssec:vin_training}. After evaluating all $n$ query views, the VIN-NBV policy greedily selects the one with the highest improvement score. We move to the selected view position $C^t_*$ and acquire new RGB-D capture which we use to update $I^t_{base}$ $C^t_{base}$ $D^t_{base}$ to get $I^{t+1}_{base}$ $C^{t+1}_{base}$ $D^{t+1}_{base}$. Once we have reached our desired stopping criteria, we create and return our final 3D reconstruction $R_{final}$ using $I^{m}_{base}$ $C^{m}_{base}$ $D^{m}_{base}$.

Although we adopt a simple greedy acquisition strategy, our policy provides the flexibility to incorporate any desired constraints into the agent's acquisition strategy. In Section~\ref{sec:results}, we evaluate the VIN-NBV policy with constraints on the number of acquisitions and time in motion for the capture agent. Our method also allows the user to specify any sampling strategy to create a set of query views to evaluate, providing high adaptability to downstream applications.

\subsection{Design of VIN}\label{ssec:vin_design}
\vspace{-0.25em}

The View Introspection Network (VIN) estimates the utility of a candidate viewpoint by predicting its Relative Reconstruction Improvement (RRI) given the acquisition history and current reconstruction. There are three stages: first it reconstructs the scene, then it featurizes its geometry, and finally it encodes features to predict improvement.

From the current set of base RGB-D images $I_{base}$ and camera parameters $C_{base}$, we reconstruct a 3D point cloud $\mathcal{R}_{base}$. If a depth sensor is unavailable, monocular depth estimation or Multi-View Stereo can be substituted. Each point in $\mathcal{R}_{base}$ is enriched with surface normals, visibility count, and depth values. Surface normals estimate local geometry, with variance across neighboring points indicating complex surfaces that may require additional captures; low variance indicates planar regions that need less captures. Visibility count tracks the number of views from $I_{base}$ in which a point is observed. Points frequently seen in many views are less informative for new captures; points rarely seen may be more informative. Depth values provide distance information for projected pixels, helping to detect surface inconsistencies and missing regions.

For each candidate view $C_q$ the enriched point cloud is projected into its image plane, forming a feature grid (size 512$\times$512$\times$5). Here each pixel contains surface normals, visibility count, and depth after projection. We downsample this grid to 256$\times$256 via pooling to reduce computation and compute per-pixel variance to capture local geometric complexity. The down sampled feature grid is represented as $F_p$ and the per-pixel variances are $F_v$.

We also compute an empty feature $F_{empty}$ to provide explicit information about reconstruction coverage. This allows the VIN to focus on learning key information on top of coverage to predict reconstruction improvement. We consider a pixel to be "empty" if no point in $\mathcal{R}_{base}$ projects to it. We form a hull around non-empty pixels and count empty pixels inside and outside of the hull. Empty pixels inside the hull expose “holes” in the existing reconstruction and empty pixels outside the hull indicate potentially unseen geometry. We concatenate both of these values to create the two-element $F_{empty}$ feature vector. We use $F_{empty}$ as a lightweight proxy for more traditional reconstruction coverage measures.

To encode the features we define a convolutional encoder $V_{view} = \mathcal{E}_\theta\bigl(F_v \oplus F_p\bigr)$ that is applied to the feature grids and variances to transform the local per-pixel information into a global view feature vector $V_{view}$ of size $256\times1$, where $\oplus$ means concatenation. In addition to $V_{view}$, we also provide the number of base views $F_{base}$ to help indicate at what stage of the capture we are in; since helpful views in the earlier stages look different from helpful views in later stages, this may change how the model scores different views. We concatenate all of this information with $V_{view}$ and pass it through a MLP $\mathcal{M}(\cdot)$ to predict the final improvement score:

\vspace{-1.2em}
\begin{equation}
\widehat{\mathcal{RRI}}(q) = \mathcal{M}_\phi\bigl( \mathcal{E}_\theta\bigl(F_v \oplus F_p\bigr) \oplus F_{empty} \oplus F_{base}).
\end{equation}

\subsection{Training of VIN} \label{ssec:vin_training}
\vspace{-0.25em}

Our objective is to train the VIN to estimate the fitness of a query view, defined as its Relative Reconstruction Improvement ($\mathcal{RRI}(q)$) over the existing base views (as described in Equation \ref{eq:rri}). To achieve this, we adopt an imitation learning approach. For each object we sample and render a set of 120 RGB-D views that surround it, then at each step we
exhaustively compute the $\mathcal{RRI}$ for every query view. At each step we reconstruct the point cloud using the new image along with the previously captured base images. This computed score, which relies on access to the ground-truth 3D model and rendering engine, is referred to as the Oracle RRI. VIN is then trained to predict this Oracle RRI, without access to the rendered image, using only the RGB-D images from previously captured views and the camera parameters of the query view.

Directly regressing the $\mathcal{RRI}$ proves to be challenging. The model struggles to generalize across unseen objects and categories. To address this, we reformulate the task as a classification problem by discretizing the $\mathcal{RRI}$ into 15 ordinal classes, where class 0 indicates the least improvement and class 14 the highest. In this formulation, we recognize that misclassifications between distant classes are more detrimental than those between nearby ones. Therefore, even if VIN cannot always predict the optimal next-best view, it should still identify a reasonably good one. To enforce this, we use a ranking-aware classification loss: CORAL \cite{coral2020}. This loss transforms ordinal labels into a series of binary classification tasks, encouraging predictions that respect the natural order of the labels. This improves the model predictions and reduces large misclassifications.

An additional challenge arises in defining consistent class labels across different stages of acquisition. Early acquisitions often have higher $\mathcal{RRI}$ values because large portions of the scene remain unseen, so view selection has a greater impact. In contrast, later stages typically yield lower $\mathcal{RRI}$ values, as improvements become more incremental, focusing on resolving occlusions and small gaps. Using a fixed class assignment strategy across all stages would misclassify even the best views in later stages as lower-quality.

To overcome this, we normalize $\mathcal{RRI}$ values in a stage-independent manner. We group our training data by capture stage, defined by the number of base views, and determine the standard deviation and mean of the $\mathcal{RRI}$ values in these groups. We then take $\mathcal{RRI}$ for each query view and convert it into a z-score based on the number of standard deviations they are from the mean of their group. We soft-clip the z-scores using a tanh function, to prevent extreme outliers, and then group views into 15 dynamically sized bins, ensuring a similar number of samples in each bin. We use these bins as our final class labels.
\section{Experiments}
\label{sec:results}
\vspace{-0.25em}

\begin{figure}
  \centering
    \includegraphics[width=\columnwidth]{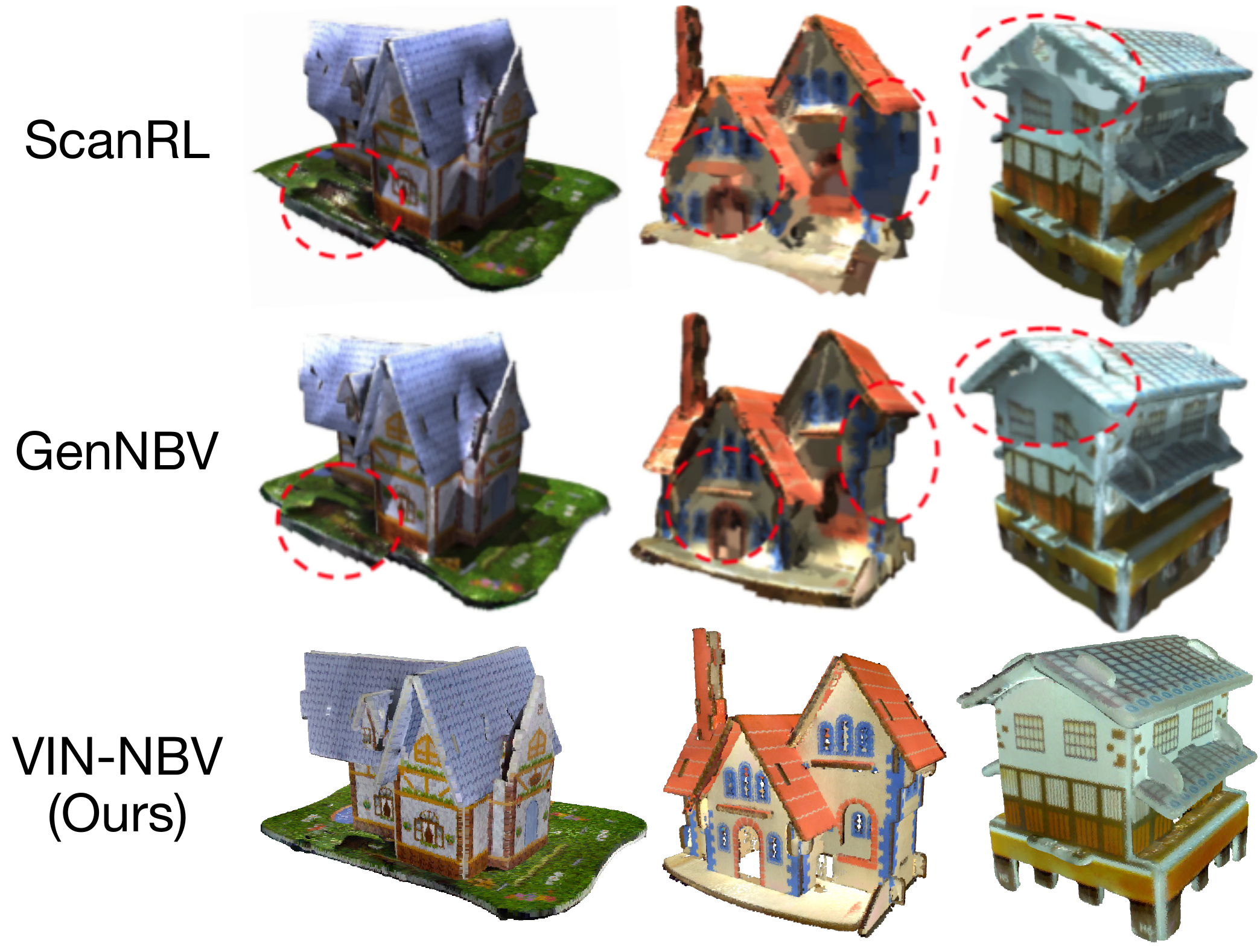}
    \vspace{-0.5em}
    \caption{We compare reconstruction quality of our VIN-NBV policy with GenNBV \cite{chen2024gennbv} and ScanRL \cite{peralta2020next} on 3 objects using 20 captures and a modified version of the figure also depicted in the original GenNBV \cite{chen2024gennbv} paper.}
    \label{fig:three_house}
    \vspace{-1em}
\end{figure}

\begin{table}
\vspace{0.75em}
  \centering
    \small
    \begin{tabular}{@{}lc@{}}
      \toprule
      \multicolumn{2}{c}{\textbf{OmniObject3D}} \\
      \midrule
      \textbf{NBV Policy} & Chamfer Distance (cm) $\downarrow$ \\
      \midrule
      Random Hemisphere   & 0.48 \\
      Uniform Hemisphere  & 0.41 \\
      \midrule
      Uncertainty-Guided  & 0.41 \\
      ActiveRMAP, Arxiv'22\cite{zhan2022activermapradiancefieldactive}          & 0.38 \\
      \midrule
      Scan-RL, ECCV'20\cite{peralta2020next} & 0.37 \\
      GenNBV, CVPR'24\cite{chen2024gennbv} & 0.33 \\
      \textbf{VIN-NBV (Ours)}      & \textbf{0.20} \\
      \bottomrule
    \end{tabular}
    \vspace{-0.25em}
    \caption{Quantitative evaluation of NBV policies on OmniObject3D \cite{wu2023omniobject3d} houses (20 acquisitions) using Chamfer Distance. We follow the evaluation setup of GenNBV and report the performance of existing methods from their paper.}
    \label{table:result_chamfer_distance}
  \vspace{0.5em}
\end{table}

We evaluate and compare VIN-NBV to existing NBV approaches that select viewpoints for high-quality 3D reconstructions on standard datasets. We focus on scenarios where an agent can acquire only a few images or has a limited motion time.

\noindent\textbf{Datasets.} We follow the same train-test protocol from GenNBV\cite{chen2024gennbv} by training on a modified subset of Houses3K \cite{peralta2020next}, and testing on the house category from OmniObject3D \cite{wu2023omniobject3d}. We also evaluate on the dinosaur, truck, and animal classes from OmniObject3D \cite{wu2023omniobject3d} to show the generalization beyond the training data (see Section \ref{sec:generalization}). Each object is rendered from 120 views and we begin the acquisition process by selecting the first base view randomly and then set the second base view to be the one closest to the first.

\noindent\textbf{Baselines.} Prior NBV approaches explicitly "look ahead” from hypothetical camera positions by evaluating visibility of the current scene model, either volumetrically by ray-casting through an occupancy/TSDF grid to estimate information gain \cite{Islerinformationgainvolumetric3d, hardouinUAVNBV}, or via rendering-style approximations of coverage from a discrete set of viewpoints \cite{connollydetermination}, with related ray/visibility accounting in sensor-aware formulations \cite{pito1999solution} and quality-driven surface methods \cite{wuqualitydriven}. Following this approach, we design the Coverage-NBV (Cov-NBV) baseline that scores a candidate view by projecting the current 3D reconstruction into its image plane. 

Since our primary goal is to show that our proposed Relative Reconstruction Improvement (RRI) criterion, predicted by the View Introspection Network (VIN), is better than the coverage-based criterion, we use the same sampling-based sequential greedy best next view selection strategy for both VIN-NBV and Cov-NBV. We simply replace the $\mathcal{RRI}$ score (line~\ref{line:score_func} predicted by VIN in Algo.~\ref{method_nbv_algorithm}), with a coverage-based fitness function eq. \ref{eq:cov-fitness1} where $W\times H$ is the image resolution and $\mathbbm{1}(\cdot)$ is an indicator function.

\vspace{-1.0em}
\begin{equation}
Cov(q) =W\times H - \sum_{u=1}^{H} \sum_{v=1}^{W} \mathbbm{1}(C_q(\mathcal{R}_{base})_{u,v})
\label{eq:cov-fitness1}
\end{equation}
\vspace{-0.5em}

This function estimates the number of empty pixels in the query view $C_q$ when rendering the current reconstructed 3D point cloud $\mathcal{R}_{base}$. A higher $Cov(q)$ score suggests that the query view covers more previously unseen areas, making it a potentially valuable viewpoint.
This pixel-space surrogate correlates with reconstruction completeness and serves as a strong, transparent baseline. It is sensor-agnostic, cheap to compute with standard z-buffer rendering, and captures the same “look-ahead visibility” principle that underlies volumetric information-gain methods—without requiring full voxel ray marching.

We also compare our VIN-NBV policy against several existing 3D-based next-best-view (NBV) approaches, including ScanRL \cite{peralta2020next}, GenNBV \cite{chen2024gennbv}, and ActiveRMAP \cite{zhan2022activermapradiancefieldactive}. Since the model weights of GenNBV\cite{chen2024gennbv} are unavailable, we compare with their reported results in the paper directly. We also perform a direct visual comparison to the reconstruction results provided in their paper, as seen in Fig.~\ref{fig:three_house}.

To understand the upper bound performance of our methods in the optimal case we propose Oracle NBV: a variant of the VIN-NBV policy that selects the best image at each step where the fitness function (line~\ref{line:score_func} in Algorithm~\ref{method_nbv_algorithm}) is replaced with the ground-truth Relative Reconstruction Improvement ($\mathcal{RRI}$) of each query view. This method assumes access to the complete ground-truth reconstruction, enabling direct computation of $\mathcal{RRI}$ as defined in Eq.~\ref{eq:rri}. 

For all sampling-based acquisition policies, VIN-NBV, Cov-NBV, and Oracle-NBV, we uniformly render 120 viewpoints in 3 hemispherical shells around the object from which the policies can sample during evaluation.

\noindent\textbf{Metrics.} Since our goal is to improve reconstruction quality, we use the Chamfer Distance as our main accuracy metric, as it can capture fine-grained information at the point level. We also include the coverage percentage and F1 score curves to show the change in reconstruction completeness and quality as more acquisitions are made. For each object, we calculate the metrics between the reconstructed and the ground truth point clouds. We report the average accuracy (Chamfer Distance) in centimeters across all objects in Omniobject3D houses \cite{wu2023omniobject3d}.

\begin{figure}
  \centering
  \includegraphics[width=0.96\columnwidth]{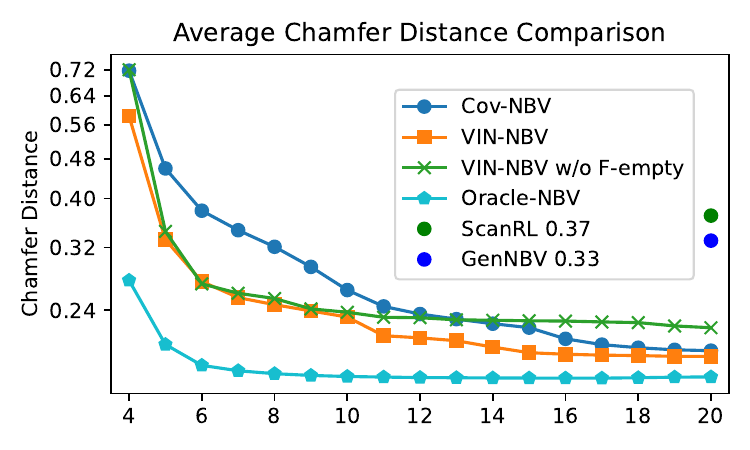}
    \vspace{-1em}
    \caption{
    \textbf{Evaluation with Limited Acquisitions.} Average reconstruction error (in cm) on OmniObject3D \cite{wu2023omniobject3d} houses under constraint on number of acquisitions. VIN-NBV consistently outperforms coverage-based NBV policies.
    }
    \label{fig:chamfer_comparison}
    \vspace{0.5em}
\end{figure}

\subsection{Evaluation with Limited Acquisitions}
\label{limited_capture_results}
\vspace{-0.25em}

To enable direct comparison with GenNBV and ScanRL \cite{qi2023my3dgen, peralta2020next}, we evaluate our policy on the OmniObject3D \cite{wu2023omniobject3d} houses dataset, limited to 20 captures. VIN-NBV outperforms state-of-the-art NBV policies (Table \ref{table:result_chamfer_distance}), achieving an average reconstruction error of 0.20 cm on houses, compared to 0.33 cm for GenNBV \cite{chen2024gennbv} and 0.37 cm for ScanRL. A visual comparison is presented in Fig. \ref{fig:three_house}, modified directly from the final reconstructions presented in the GenNBV \cite{chen2024gennbv} paper, and shows VIN-NBV retains fine detail and overall good reconstruction quality whereas GenNBV and ScanRL produce blurrier results despite good coverage, likely due to the authors' use of mesh renderings as opposed to dense clouds.

In Fig.~\ref{fig:chamfer_comparison} we compare the reconstruction accuracy during the intermediate stages of acquisition between VIN-NBV with the coverage baseline 'Cov-NBV'. We observe that large improvements with respect to the coverage baseline happen during the initial stages of capture, while towards the later stages, both policies result in similar reconstruction quality. Since the model weights and the evaluation code of the GenNBV \cite{chen2024gennbv} paper are unavailable, we were unable to evaluate this method during the intermediate stages of the capture process. The gap between the 'Oracle NBV' and our method shows that there is still room for improvement, especially in the early stages.

\begin{figure}
  \centering
  \begin{subfigure}[b]{0.49\columnwidth}
    \centering
    \includegraphics[width=\linewidth]{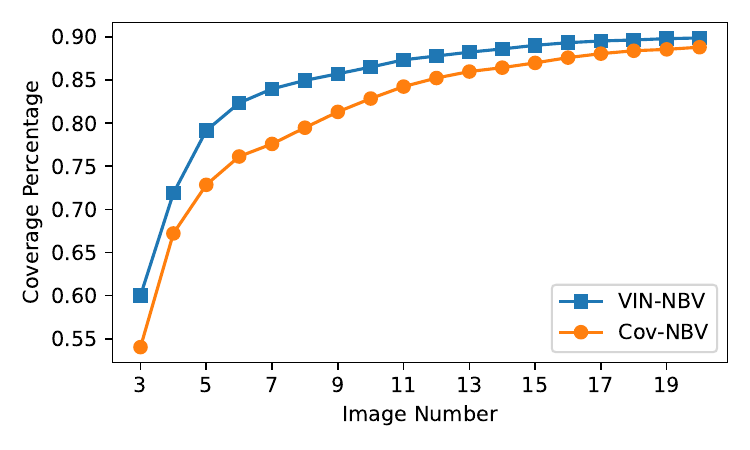}
    \caption{Coverage \% over captures}
    \label{fig:house_coverage_curve}
  \end{subfigure}\hfill
  \begin{subfigure}[b]{0.49\columnwidth}
    \centering
    \includegraphics[width=\linewidth]{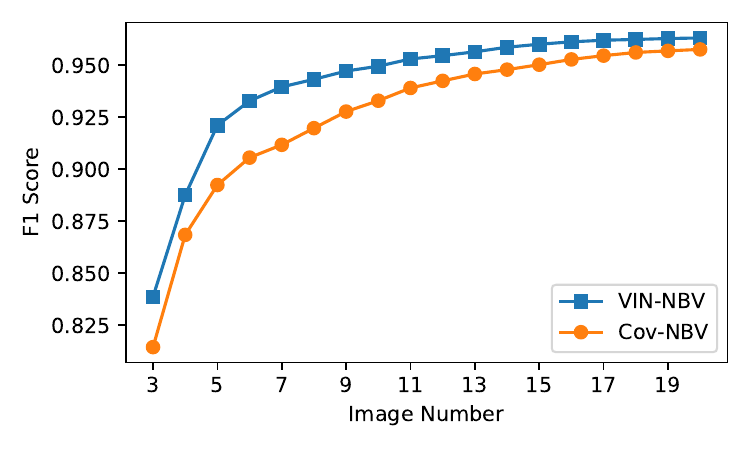}
    \caption{F1 score over captures}
    \label{fig:house_fcr_bar}
  \end{subfigure}
  \vspace{-0.5em}
  \caption{%
    \subref{fig:house_coverage_curve} mean coverage percentage and \subref{fig:house_fcr_bar} F1 score across all OmniObject3D house objects at each acquisition step.
  }
  \label{fig:coverage_metrics_two_panel}
  \vspace{0.5em}
\end{figure}

\noindent\textbf{Ablation: Coverage feature $F_{empty}$ in VIN.} To understand the importance of the coverage information we provide with $F_{empty}$, we run an ablation that evaluates the VIN-NBV policy using a VIN trained without $F_{empty}$. In Fig. \ref{fig:chamfer_comparison}, we see that the absence of coverage features makes little difference during the earlier acquisition stages, but is substantial in the later ones. This explains why VIN-NBV achieves large gains over Cov-NBV early on, converging to a similar solution during later stages where coverage is helpful. Although VIN focuses on predicting reconstruction quality, the coverage feature is still a helpful indicator,  which is otherwise hard to learn implicitly during training.

\subsection{Evaluation with Time-Limited Agent Motion}
\label{limited_time_results}
\vspace{-0.25em}

In our time-limited setting, we evaluate our policy under varying constraints on time in motion to understand its efficiency under strict budgets, simulating time-sensitive applications. We assume that the agent is a drone equipped with a depth sensor and camera and travels at a constant velocity of 4 mph, a typical speed for consumer-grade drones. We assume that the drone always takes a straight-line path to the next capture position and determine how far it can travel under different time limits. We use the same OmniObject3D \cite{wu2023omniobject3d} evaluation data as before and scaled all objects to a size of 14 meters to simulate a more realistic capture setting.

\noindent\textbf{Adapting NBV policies for time-limited acquisitions.} At each step during capture, our agent checks how far it can still travel and finds all potential next views that are in range. If our agent is not able to find a next view in range, it stops capturing and computes the Chamfer Distance of the reconstruction with the current captures. We use our VIN-NBV policy to select the NBVs and don't impose a limit on the number of images that the agent can capture.

\begin{figure}
  \centering
  \vspace{-1em}
  \includegraphics[width=0.94\columnwidth]{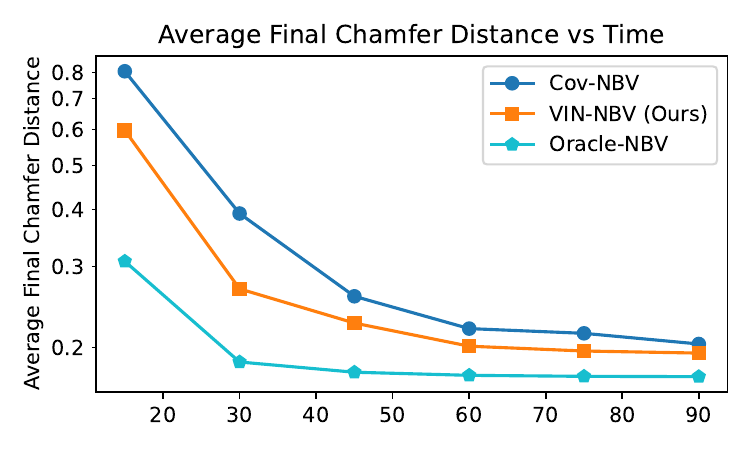}
    \vspace{-1em}
    \caption{
    \textbf{Evaluation with Time-Limited Agent Motion.} Average reconstruction error (in cm) on OmniObject3D \cite{wu2023omniobject3d} houses under constraint on time traversed. VIN-NBV consistently outperforms coverage based NBV policies.
    }
    \label{fig:time_constraint}
    \vspace{0.5em}
\end{figure}

\begin{figure*}[!t]
  \centering
  \vspace{0.25em}
\includegraphics[width=\textwidth]{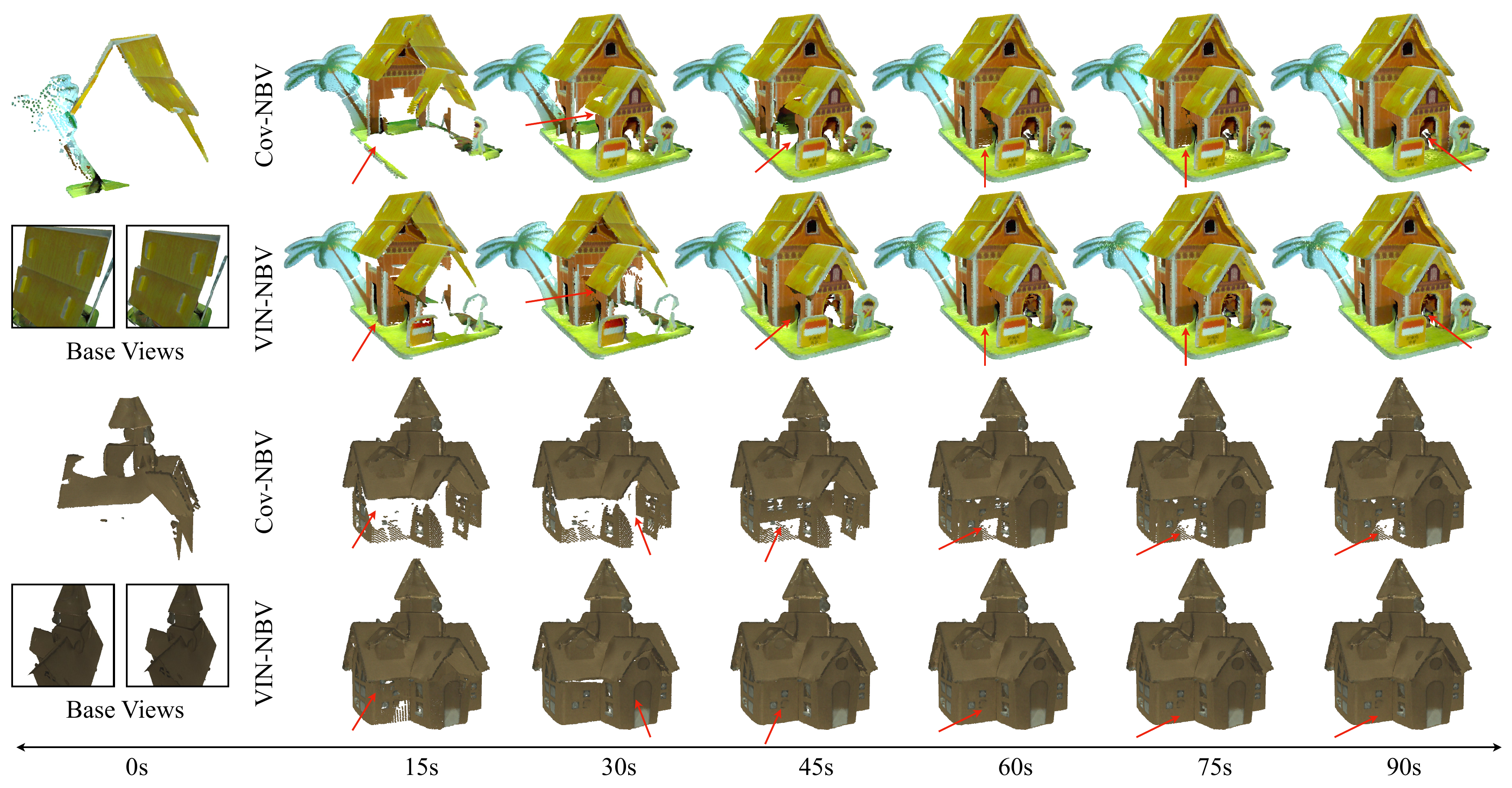}
    \vspace{-1.5em}
  \caption{We visualize reconstruction obtained by VIN-NBV and Cov-NBV for different constraints on time traversed. For smaller time budgets, VIN-NBV reconstructions are better with fewer holes.}
  \label{fig:time_figs}
  \vspace{-1.0em}
\end{figure*}

We show results from the time-limited setting in Fig. ~\ref{fig:time_constraint}. Under all time constraints, our method outperforms the coverage baseline, with the largest improvement happening under the strictest time constraints. During the 15-second limit, our method beats the coverage baseline Chamfer Distance by $\sim$25\%. In Fig.~\ref{fig:time_figs}, we visualize the reconstructed 3D scene using our VIN-NBV policy and the coverage baseline (Cov-NBV) for varying time constraints, illustrating that with more time, the policy focuses on capturing views that can fill missing regions and holes in the reconstruction.

\begin{figure}
  \centering
  \vspace{-1.3em}
\includegraphics[width=0.94\columnwidth]{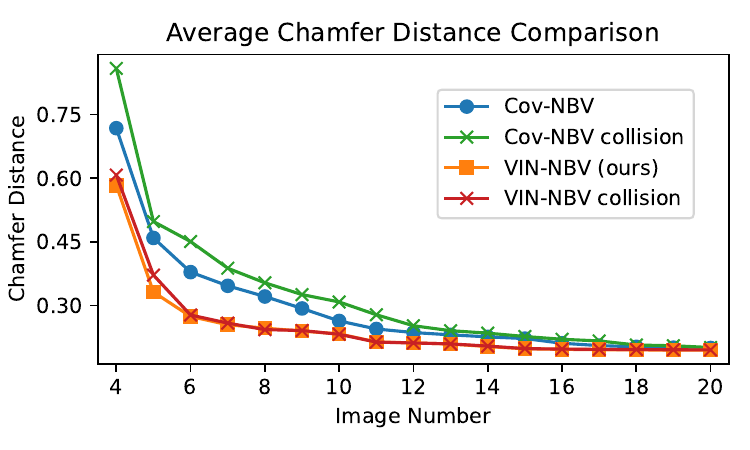}
  \vspace{-1.3em}
    \caption{
        \textbf{Evaluation with Collision Avoidance.} Comparison of average reconstruction error (in cm) on OmniObject3D \cite{wu2023omniobject3d} houses with collisions avoidance planning. VIN-NBV performance is similar while Cov-NBV performance decreases.
    }
    \label{fig:collision_comparison}
    \vspace{0.5em}
\end{figure}

\subsection{Collision Avoidance}
\label{sec:collision_avoidance}
\vspace{-0.25em}

Currently our capture agent assumes that it can perform straight line travel between the current position and the selected next view. This approach does not take into account collisions that may occur from following these straight line trajectories. We demonstrate the ability of our method to handle custom collision criteria by running the evaluation in a setting that requires the selected NBV to have a straight line path that is at least 1 meter away from the reconstruction at all times. In Fig.~\ref{fig:collision_comparison}, we see that even with this additional collision criterion integrated into our NBV-policy, it still performs well and outperforms the Cov-NBV baseline.

\begin{figure*}[!t]
  \centering
  \vspace{-1.0em}
  \begin{subfigure}[t]{0.33\textwidth}
    \centering
    \includegraphics[width=\textwidth]{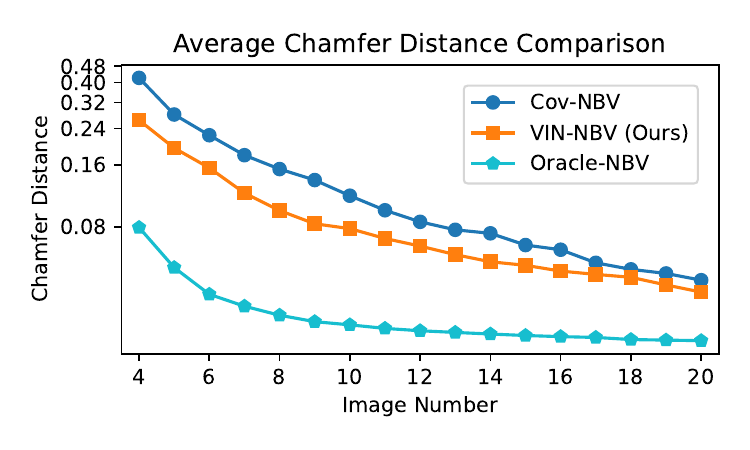}
    \caption{Dinosaurs}
    \label{fig:dinosaur_chamf}
  \end{subfigure}
  \hfill
  \begin{subfigure}[t]{0.33\textwidth}
    \centering
    \includegraphics[width=\textwidth]{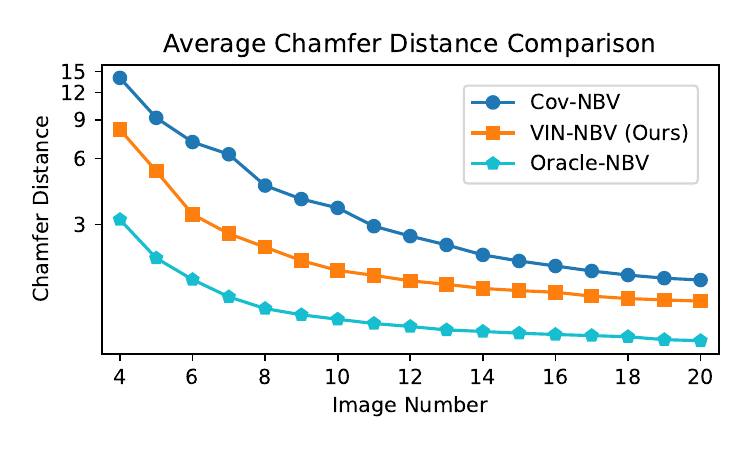}
    \caption{Toy Motorcycles}
    \label{fig:motorcycle_chamf}
  \end{subfigure}
  \hfill
  \begin{subfigure}[t]{0.33\textwidth}
    \centering
    \includegraphics[width=\textwidth]{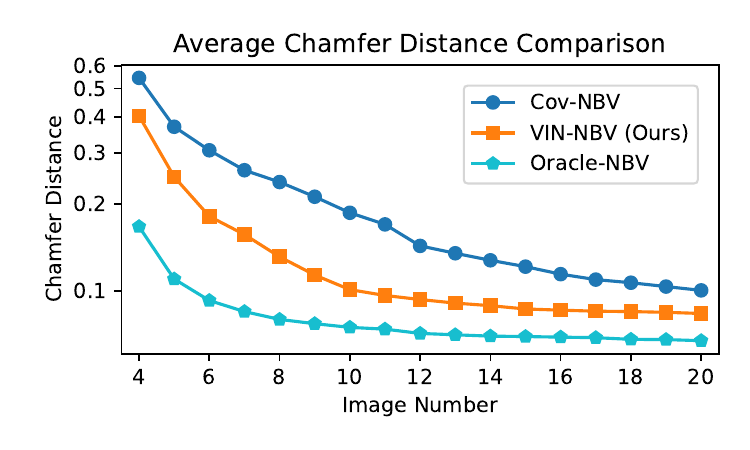}
    \caption{Toy Animals}
    \label{fig:animal_chamf}
  \end{subfigure}
  \vspace{-0.5em}
  \caption{
  \textbf{Generalization Across Object Categories.} We graph the average Chamfer Distance at each capture stage in three additional object classes namely dinosaurs, toy motorcycles,
  and toy animals. We compare VIN-NBV to a coverage baseline for 20 captures.
  }
  \label{fig:all_generalization}
\end{figure*}

\subsection{Generalization Across Object Categories}
\label{sec:generalization}
\vspace{-0.25em}

We evaluate our policy on additional object categories from OmniObject3D \cite{wu2023omniobject3d}, namely dinosaurs, toy animals, and toy motorcycles. Many of the house objects contain largely planar surfaces with large flat regions and overall less curvature and self-occlusion, in contrast the additional object classes we evaluated have much more detailed surface geometry and instances of self-occlusion. Our results show that even with these new complexities our policy is able to generalize well.

In Fig.\ref{fig:all_generalization} we see that when we evaluate our VIN-NBV policy on additional object categories, it is able to consistently outperform the coverage baseline. Larger gaps in performance occur early on, with both methods reaching similar final Chamfer Distance values at 20 captures. Our policy does particularly well, compared to the coverage baseline, for objects with more complex self-occlusions such as toy motorcycles. We include visual results of our policy and the coverage baseline after 10 acquisitions on several objects and from various viewing angles in Appendix \ref{sec:additional_visuals}.

\begin{table}
\vspace{0.5em}
    \centering
    \begin{tabular}{lccc}
    \toprule
    \textbf{Policy}   & \textbf{Dinosaurs} & \textbf{Trucks} & \textbf{Animals} \\
    \midrule
    Scan-RL \cite{peralta2020next}        & 0.06 & 0.88 & 0.17 \\
    GenNBV \cite{chen2024gennbv}         & \textbf{0.03} & 0.81 & 0.16 \\
    VIN-NBV (ours)       & 0.04   & \textbf{0.57}  & \textbf{0.08}   \\
    \bottomrule
    \end{tabular}
    \vspace{-0.5em}
    \caption{Reconstruction quality comparisons to prior works, with Chamfer Distances ($\downarrow$ is better), on 3 additional classes from OmniObject3D \cite{wu2023omniobject3d} after 20 total acquisitions.}
    \label{table:additional_classes_table}
    \vspace{0.5em}
\end{table}

\vspace{-0.5em}
The toy truck category was also evaluated for final Chamfer Distance so that direct comparisons against GenNBV’s \cite{chen2024gennbv} reported results could be made in Table~\ref{table:additional_classes_table}. VIN-NBV outperforms ScanRL but is slightly behind GenNBV on dinosaurs, while significantly beating both baselines on toy animals and trucks.
\section{Limitations}
\label{sec:limitations}
\vspace{-0.25em}

We use ground truth depth maps in RGB-D capture to match the evaluation setting of the prior works GenNBV \cite{chen2024gennbv} and ScanRL \cite{peralta2020next}, but these depth maps are noise-free and not reflective of true depth sensor readings. The performance of our method has not been evaluated with real world data and could have different performance in such settings. In Fig. \ref{fig:chamfer_comparison} we see a gap to the Oracle-NBV in early capture stages, indicating that our method still has room for improvement.
\section{Conclusion}
\label{sec:conclusion}
\vspace{-0.25em}

In this paper, we revisit the Next Best View problem and propose the VIN and VIN-NBV policy, a generalizable policy that can determine a set of optimal acquisitions to maximize reconstruction quality. Our policy can be easily adapted to operate with limitations on the number of acquisitions or a time limit on agent motion. By optimizing for reconstruction quality rather than traditional coverage, the VIN-NBV policy improves final reconstruction results without requiring prior scene knowledge, extra image captures, per-scene training, or complex RL policies. Evaluations on OmniObject3D \cite{wu2023omniobject3d} show VIN-NBV outperforms state-of-the-art RL coverage based methods, reducing reconstruction error by up to 40\%. The VIN-NBV policy can enable agents to efficiently acquire necessary 3D information for various applications where time is of the essence.

{
    \small
    \bibliographystyle{ieeenat_fullname}
    \bibliography{main}
}

\clearpage
\section{Appendix}
\label{sec:appendix}

\subsection{Overview}
\noindent Our appendix includes the following:
\begin{itemize}
  \item \textbf{Section~\ref{sec:implementation}.} More details on model implementation and data setup.
  \item \textbf{Section~\ref{sec:additional_visuals}.} The chamfer distance graph for the truck category from OmniObject3D \cite{wu2023omniobject3d} and various visual results of our policy and the coverage baseline after 10 acquisitions on several objects and from various viewing angles. 
\end{itemize}

\subsection{Implementation Details}
\label{sec:implementation}

\noindent\textbf{Model Implementation.} We implement our model using Pytorch \cite{paszke2019pytorchimperativestylehighperformance} 
and carry out training using Pytorch Lightning \cite{Falcon_PyTorch_Lightning_2019}.
We leverage Pytorch3D\cite{ravi2020pytorch3d} to help create and render our point clouds. Our convolutional encoder has 4 layers and hidden dimension size of 256. Our rank MLP has 3 layers with a hidden dimension size of 256 and uses a CORAL \cite{coral2020} layer as its final layer so that a CORAL loss can be used during training. We use the open-source implementation of this loss and other necessary components provided by the authors on \href{https://github.com/Raschka-research-group/coral-pytorch}{GitHub}.

\noindent\textbf{Point Cloud Projection.} To project our point clouds to query views, we use the Pytorch3D\cite{ravi2020pytorch3d} PointsRasterizer with a points per-pixel setting of 1 and a fixed radius. Since the rasterizer uses a normalized coordinate system and our voxel down sampling leads to evenly spaced points, we find that this fixed radius size is sufficient. During the projection process for the point cloud we save a mapping from pixel to point allowing us to determine per-pixel feature vectors to construct a feature grid. We do all projection in batch.

\noindent\textbf{Model Training.} We train our model for 60 epochs using the AdamW\cite{loshchilov2019decoupledweightdecayregularization} optimizer and a cosine annealing\cite{loshchilov2017sgdrstochasticgradientdescent} learning rate scheduler. For our learning rate, we used 1e-3 and train for approximately 24 hours on four A6000 GPUs. During training time, we treat each set of next views for one specific object as a single batch. Batching by object means we only need to reconstruct one point cloud for an entire batch and can project it to all candidate views in one go, saving memory and computation.

\noindent\textbf{Evaluation Data Preparation.} The authors of GenNBV \cite{chen2024gennbv} resized the house objects from OmniObject3D \cite{wu2023omniobject3d} to better fit their problem context. The resized objects were made available by the authors on the project GitHub and we used them to calculate the exact scale factor applied to the original OmniObject3D \cite{wu2023omniobject3d} houses. We use this scale factor to bring our Chamfer Distance values computed on the unscaled objects into the correct scale for direct comparison. We also note that the point cloud for house 27 is missing and thus do not include it in evaluation since the correct scale factor cannot be calculated.

\begin{figure}[H]
  \centering
  \includegraphics[width=0.96\columnwidth]{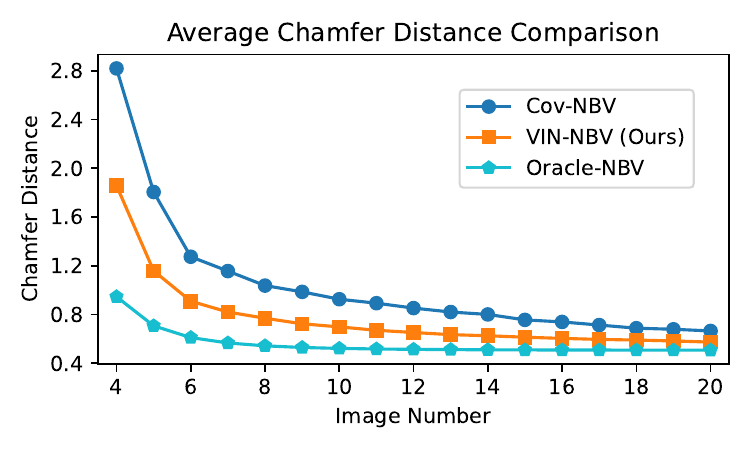}
    \caption{Average Chamfer Distance across all truck objects from OmniObject3D \cite{wu2023omniobject3d} across each acquistion step.}
    \label{fig:truck_chamfer}
    \vspace{0.5em}
\end{figure}

\subsection{Additional Visualizations}
\label{sec:additional_visuals}

\begin{figure*}
  \centering
  \includegraphics[
  page=3,width=\textwidth,
  trim=10mm 12mm 10mm 12mm,clip,
  ]{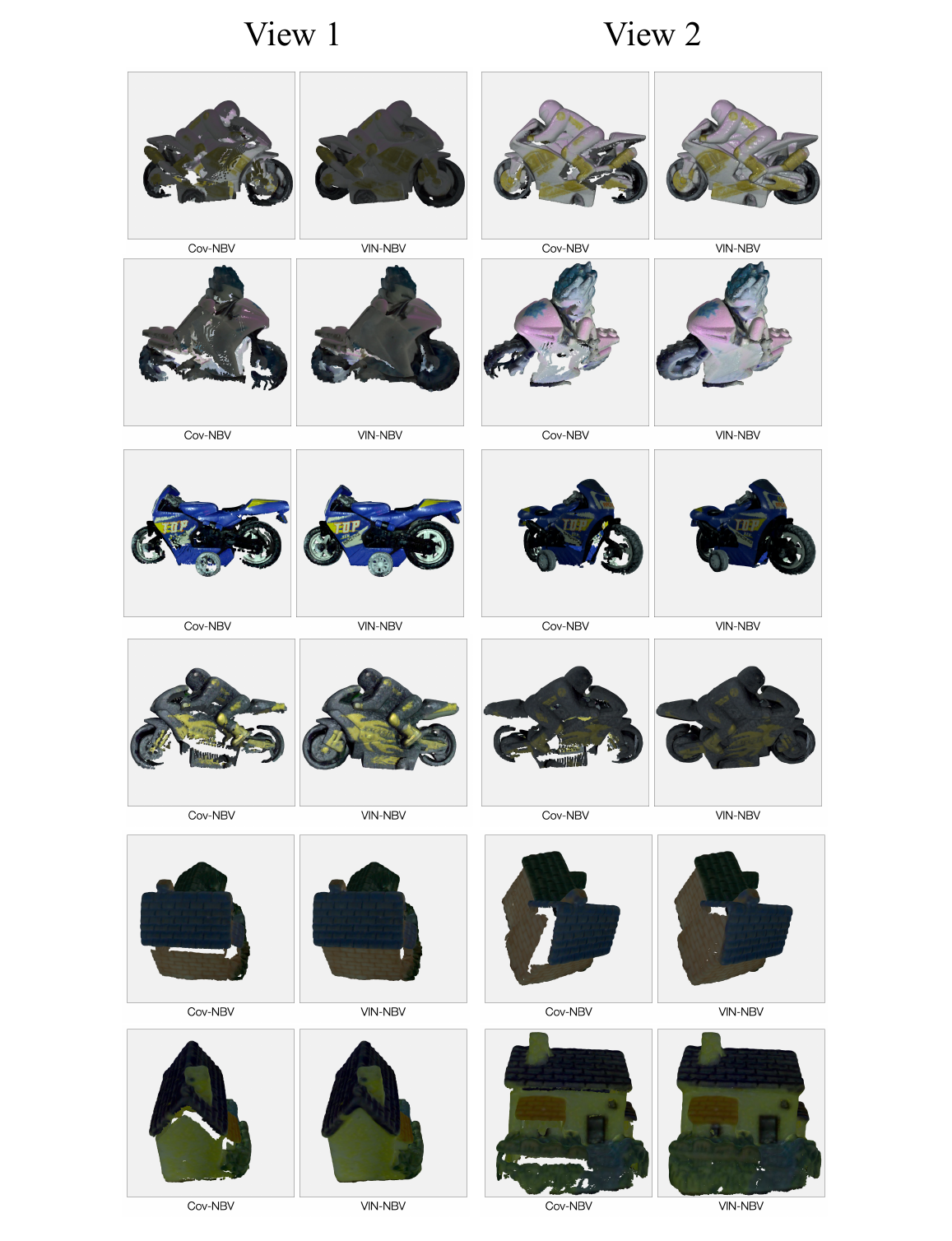}
  \vspace{-2.5em}
  \caption{
    Comparison of VIN-NBV and Cov-NBV after 10 acquisitions on motorcycles and houses from OmniObject3D\cite{wu2023omniobject3d}.
  }
  \label{fig:extra_visuals_p3}
\end{figure*}

\begin{figure*}
  \centering
  \includegraphics[
  page=2,width=\textwidth,
  trim=10mm 12mm 10mm 12mm,clip,
  ]{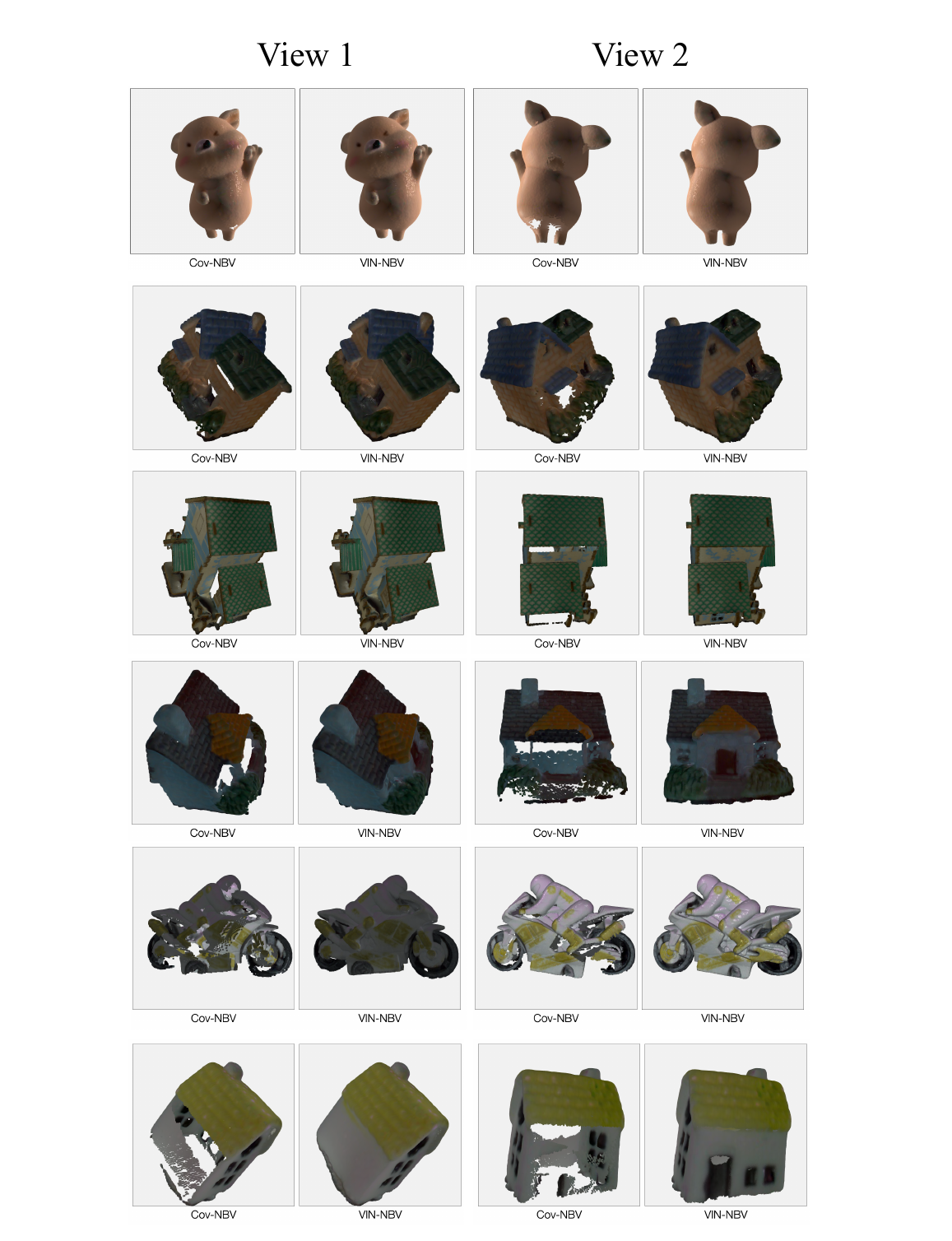}
  \vspace{-2em}
  \caption{
    Comparison of VIN-NBV and Cov-NBV after 10 acquisitions on animals, house, and motorcycles from OmniObject3D\cite{wu2023omniobject3d}.
  }
  \label{fig:extra_visuals_p2}
\end{figure*}

\begin{figure*}
  \centering
  \includegraphics[
  page=1,width=\textwidth,
  trim=10mm 12mm 10mm 12mm,clip,
  ]{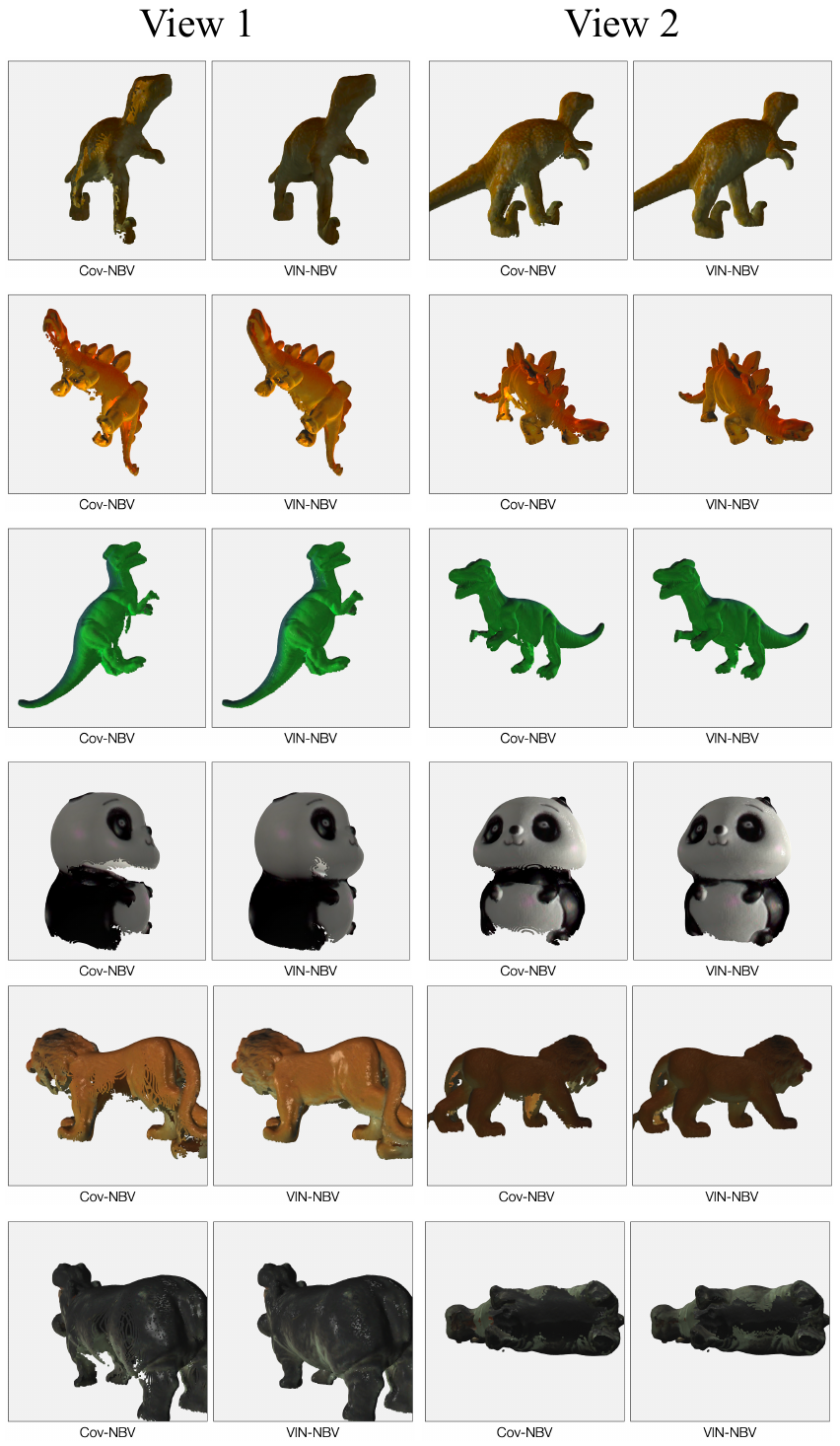}
  \vspace{-2em}
  \caption{
    Comparison of VIN-NBV and Cov-NBV after 10 acquisitions on dinosaurs and animals from OmniObject3D\cite{wu2023omniobject3d}.
    }
  \label{fig:extra_visuals_p1}
\end{figure*}

\begin{figure*}
  \centering
  \includegraphics[
  page=4,width=\textwidth,
  trim=10mm 12mm 10mm 12mm,clip,
  ]{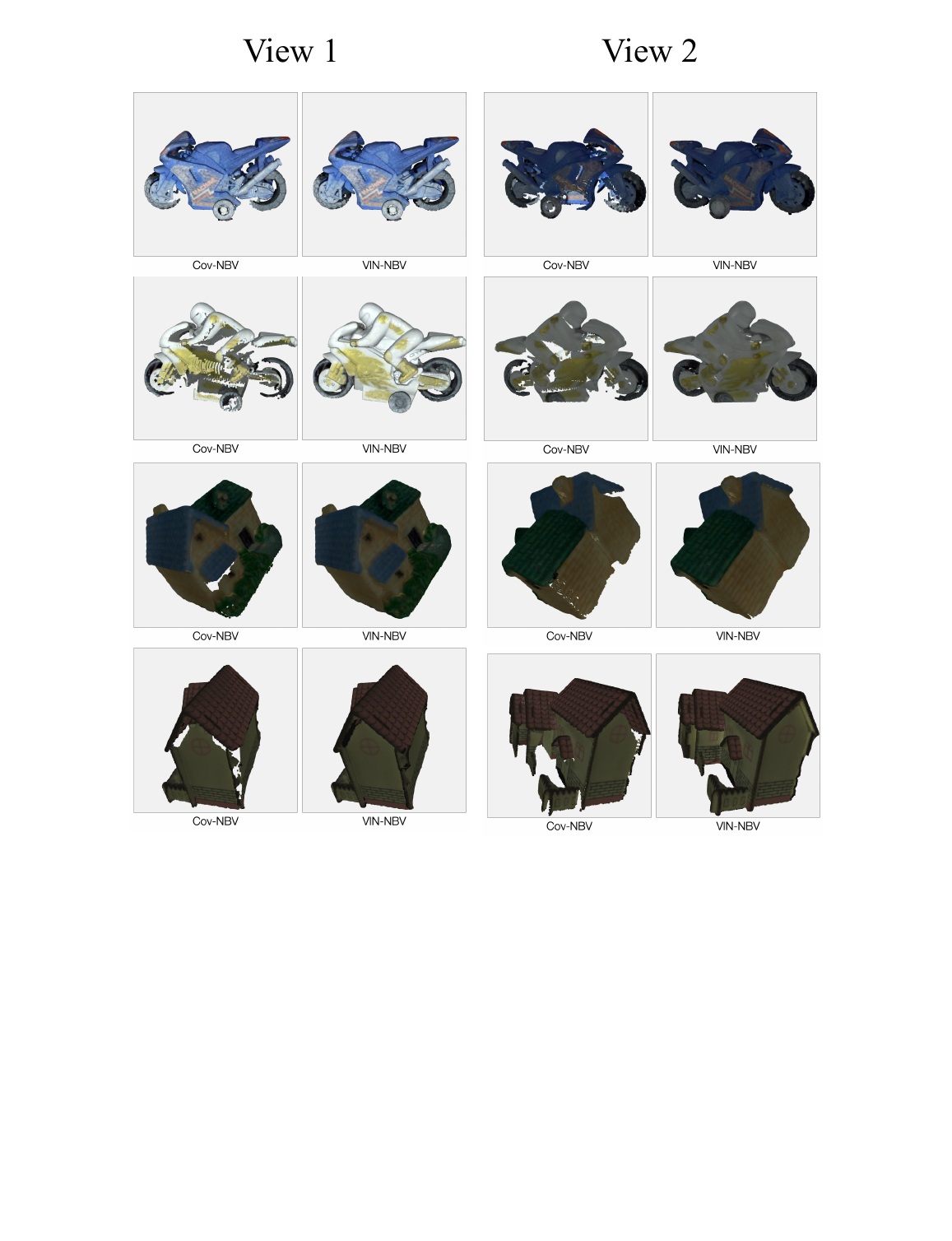}
  \vspace{-20em}
  \caption{
    Comparison of VIN-NBV and Cov-NBV after 10 acquisitions on motorcycles and houses from OmniObject3D\cite{wu2023omniobject3d}.
  }
  \label{fig:extra_visuals_p4}
\end{figure*}

We include in the following pages visualizations of various objects after acquiring 10 captures in Fig.~\ref{fig:extra_visuals_p3}, Fig.~\ref{fig:extra_visuals_p2},
Fig.~\ref{fig:extra_visuals_p1}, and
Fig.~\ref{fig:extra_visuals_p4}. We compare the results of using VIN-NBV (ours) against a simple coverage baseline Cov-NBV. We show for each object two different perspectives to provide a comprehensive overview of the final reconstruction.

\clearpage

\clearpage

\clearpage

\clearpage

\end{document}